\documentclass[12pt]{article}
\usepackage[T1]{fontenc}
\usepackage[utf8]{inputenc} 
\usepackage{lmodern}  
\usepackage[ngerman,english]{babel}
\usepackage[a4paper,margin=3cm]{geometry}

\usepackage[backend=biber,style=vancouver, maxbibnames=6, minbibnames=1]{biblatex}
\DeclareNameAlias{sortname}{family-given}
\DeclareNameAlias{default}{family-given}

\usepackage{graphicx, amsmath, amsfonts, amssymb, mathtools, subcaption, multirow, makecell, authblk, float, comment, helvet, setspace, bbm, booktabs, tabularx, array, mathrsfs, bm, upgreek, hyperref}

\title{Using latent representations to link disjoint longitudinal data for mixed-effects regression}

\author[1,2,*]{Clemens Schächter}
\author[1,2]{Maren Hackenberg}
\author[1,2]{Michelle Pfaffenlehner}
\author[3]{Félix B. Tambe-Ndonfack}
\author[3]{Thorsten Schmidt}
\author[4]{Astrid Pechmann}
\author[4]{Janbernd Kirschner}
\author[5,6]{Jan Hasenauer}
\author[1,2,7]{Harald Binder}

\affil[1]{Institute of Medical Biometry and Statistics (IMBI), Medical Faculty and Medical Center -- University of Freiburg, Germany}
\affil[2]{Freiburg Center for Data Analysis, Modeling and AI -- University of Freiburg, Germany}
\affil[3]{Department of Mathematical Stochastics, Mathematical Institute -- University of Freiburg, Germany}
\affil[4]{Department of Neuropediatrics and Muscle Disorders, Medical Faculty and Medical Center -- University of Freiburg, Germany}
\affil[5]{Bonn Center for Mathematical Life Sciences -- University of Bonn, Germany}
\affil[6]{Life and Medical Sciences (LIMES) Institute -- University of Bonn, Germany}
\affil[7]{CIBSS, Centre for Integrative Biological Signalling Studies -- University of Freiburg, Germany}
\affil[*]{Corresponding author, contact: clemens.schaechter@uniklinik-freiburg.de}
\date{\today}

\begin{document}

\maketitle 

\section*{Abstract}
\normalsize
Many rare diseases offer limited established treatment options, leading patients to switch therapies when new medications emerge. To analyze the impact of such treatment switches within the low sample size limitations of rare disease trials, it is important to use all available data sources. This, however, is complicated when the use of measurement instruments changes during the observation period, for example when instruments are adapted to specific age ranges. The resulting disjoint longitudinal data trajectories complicate the application of traditional modeling approaches like mixed-effects regression.
We tackle this by mapping observations of each instrument to an aligned low-dimensional temporal trajectory, enabling longitudinal modeling across instruments. Specifically, we employ a set of variational autoencoder architectures to embed item values into a shared latent space for each time point. Temporal disease dynamics and treatment switch effects are then captured through a mixed-effects regression model applied to latent representations. To enable statistical inference, we present a novel statistical testing approach that accounts for the joint parameter estimation of mixed-effects regression and variational autoencoders. The methodology is applied to quantify the impact of treatment switches for patients with spinal muscular atrophy. Here, our approach aligns motor performance items from different measurement instruments for mixed-effects regression and maps estimated effects back to the observed item level to quantify the treatment switch effect. Our approach allows for model selection as well as for assessing effects of treatment switching. The results highlight the potential of modeling in joint latent representations for addressing small data challenges.
\newline \
\newline
\noindent\textbf{Keywords:} Measurement instrument alignment, Variational autoencoder, Mixed-effects regression, Treatment switches, Statistical inference, Small data, Spinal muscular atrophy

\newpage
\section{Introduction}

Rare diseases often have a limited number of established treatments, leaving patients with few therapeutic options. Thus, when new medications emerge, patients often switch treatments. However, analyzing the impact of such treatment switches longitudinally is difficult since rare disease trials are characterized by small data challenges.~\cite{Small_data_rare_disease, rare_disease_1, rare_disease_2}. These include small sample sizes, heterogeneous data collected across different sites and varying observation frequencies. Additional difficulties arise when the preferred instruments to measure clinical outcomes change during the observation window. Such changes can occur from an evolving gold standard or measurement instruments that are tailored to patient characteristics that may change over time, like age or disease severity. However, using data from all available sources is important to make the most of the limited study population.
For example, in children affected by spinal muscular atrophy (SMA), symptom progression is monitored longitudinally using several specialized, age- and symptom-specific measurement instruments that evaluate different motor function skills~\cite{MUNILOFRA2025105341, RULM, HFMSE, CHOP, HINE, ALSFRS}. As patients age and their disease progresses, the preferred assessment instruments are adapted accordingly. Therefore, the longitudinal record often comprises only a subset of these instruments at a given time point.
 
Statistical approaches, such as mixed-effects regression, are widely used for modeling disease progression in longitudinal data~\cite{laird1982random}. They estimate both population-level effects (fixed effects) and subject-specific deviations (random effects) simultaneously. Applications span multiple diseases where disease progression is measured using a variety of motor function or cognitive assessments. They include neuromuscular disorders such as SMA~\cite{SMAMixedModel1, SMAMixedModel2, SMAMixedModel3, SMAMixedModel4}, as well as central nervous system disorders such as Parkinson’s disease~\cite{hanff2025sex, miller2019rate}, and multiple sclerosis~\cite{bralic2025correspondence, lin2006robust}. 
Although use cases vary, we aim to model the impact of switching between medications using a mixed-effects regression.  

However, there are two core issues with the application to disjoint data sources as encountered in SMA data. First, multidimensional measurements are usually aggregated into a single score to reduce the model complexity, which discards the information contained in individual test items~\cite{mcneish2020thinking}. Second, handling switches in measurement instruments is often circumvented by limiting analysis to time frames where a specific measurement instrument was applied. However, this does not provide generality for the whole observation window, as patients with an improving or worsening disease condition can transition to a different measurement instrument to circumvent ceiling or floor effects at the top or bottom of the test scale~\cite{MUNILOFRA2025105341}. This introduces a bias, while a reduction in sample size reduces statistical power, which already is an issue in rare disease settings to begin with. While results of different studies reporting treatment effects are usually aggregated using a meta-analysis~\cite{pascual2024efficacy, hagenacker2024effectiveness}, aligning results from different instruments is challenging due to within-patient correlation across instruments and different measurement scales.  
 
Motivated by these challenges in longitudinal data, we propose an approach that combines mixed-effects regression with an artificial neural network approach, namely variational autoencoders (VAEs)~\cite{VAE}. We leverage VAEs to incorporate multiple measurement instruments at the item level, to obtain a latent representation at each time point of longitudinally observed data, while linking the time points by a mixed-effects regression model. The aim is to facilitate modeling of treatment switch effects while providing an approach that allows for statistical inference on this joint low-dimensional latent representation.
To enable statistical testing, we introduce a bootstrap knockoff variable approach which corrects for potential biases caused by fitting the mixed-effects model within a VAE architecture.
As the mappings to and from the latent space are learnable and non-linear, the approach can also compensate for ceiling effects present in the datasets. Therefore, a continuous outcome linear regression in the latent space is sufficient. To reflect different aspects of the patient characteristics measured by different instruments and allow for a flexible embedding, we employ a latent space with several dimensions, and correspondingly consider mixed models with a multivariate outcome.

The problem of extracting similar information from different measurement instruments has been considered to some extent in the field of item response theory~\cite{Kolen2014}, which is typically based on factor analysis~\cite{Wirth2007, Skrondal2004}. This includes linear and non-linear factor models~\cite{Bauer2009}, and more general latent variable models~\cite{Kolen2004, Buuren2005, Heuvel2020}. Yet, these often rely on strong distributional assumptions and require anchor items present across different instruments~\cite{Heuvel2020}, which limits applicability in settings with several clinical measurement instruments.
The concept of domain adaptation, developed in the machine learning community, formalizes the challenge of linking data from a source and target domain under the assumption of a shared latent space. Applications have been particularly successful for image data~\cite{Csurka2017, Ganin2015, Guan2022}, such as in MR imaging~\cite{Goetz2016, Ghafoorian2017} or microscopy~\cite{Becker2015}. There, deep learning approaches are employed to learn domain-invariant representations~\cite{Chen2012, Long2015, Tzeng2017}. These approaches generally assume a rather large number of observations. Also, they typically do not consider time structure, and when they do, focus on time series with a large number of time points. Therefore, while the general concept of domain adaptation might be useful for longitudinal rare disease registries, there is no readily applicable approach for small numbers of patients with limited observations.
In particular, a combination of domain adaptation with statistical models for longitudinal models, such as mixed-effects regression is missing so far. While VAEs have been combined with mixed-effects models in the context of imaging data~\cite{Couronne2021, VAEMixedmodel1} and intensive care unit data~\cite{VAEMixedmodel2}, this does not cover changes between measurement instruments. In our own work, we have previously combined VAEs with ordinary differential equations (ODEs) for modeling disease trajectories in a lower-dimensional representation~\cite{Kober2022, Hackenberg2023, HacHarPfa2022}, based on a neural differential equation framework that allows for simultaneously fitting neural networks and dynamic models~\cite{Rubanova2019, Fortuin2020, Alaa2022}. While we have extended this to also allow for two different measurement instruments~\cite{instruments}, the focus on ODEs as deterministic models did not allow for statistical inference so far, as needed for statistically testing the effect of treatment switches. This could be provided by combining mixed-effects regression within VAEs that align different measurement instruments.

The manuscript is structured as follows: First, we introduce the approach in Section~\ref{methods}, after a brief overview of VAEs and multivariate linear mixed-effects models. To enable statistical testing, we introduce a bootstrap knockoff variable approach which characterizes and corrects for potential biases caused by fitting the mixed-effects model on VAE obtained latent variables. This is accompanied by an approach for quantifying effects in the latent space by mapping them back to the item representation. In Section \ref{results}, we assess the approach using data for SMA patients from the SMArtCARE registry~\cite{SMArtCARE}. This registry reports observations collected using five different motor performance measurement instruments, which need to be integrated to analyze the impact of treatment switches. We compare the performance of the proposed approach to a method based on a meta-analysis over per-measurement instrument models. Additionally, we validate the approach by evaluating results when recovering artificially added treatment switches and illustrate how to perform model selection for the latent mixed-effects model. A discussion is provided in Section \ref{discussion}, including a perspective on the potential of combining latent representations with statistical modeling for jointly analyzing data in small data settings. Finally, details on the hyperparameters, dataset and model choice are given in Section \ref{params}.

\section{Methods}
\label{methods}

In the following, we denote random variables as uppercase letters (e.g., $Y, Z$), their realizations as bold lowercase letters (e.g., $\mathbf{y}, \mathbf{z}$), their corresponding (conditional) probability distributions as uppercase $P$ with appropriate subscripts (e.g., $P_Y, P_{Z\mid \mathbf{y}}$), and their associated probability density functions as lowercase $p$ (e.g., $p_Y, p_{Z\mid \mathbf{y}}$). Matrices containing numerous observations are written in uppercase letters (e.g., $\mathbf{Y}, \mathbf{Z}$). Table \ref{tab:notation} lists the notation used throughout the manuscript.

\subsection{Variational autoencoders}

In this study, we use a set of variational autoencoder (VAE)~\cite{VAE} architectures to map data from the observation spaces of multiple measurement instruments into a shared low-dimensional representation of patient characteristics and reconstruct them back to the original space. A variational autoencoder is a generative deep learning approach, i.e., a model that allows for sampling, that learns latent representations of high-dimensional data based on the principle of variational inference~\cite{Blei_2017}. Therefore, a VAE provides tractable approximations to probability distributions without requiring restrictive distributional assumptions. In a first step, we introduce the classical VAE framework on a dataset containing i.i.d. observations from a single measurement instrument and time point, while we outline our approach to align multiple instruments and link them longitudinally via a mixed model in subsection \ref{multiple_VAEs}.

Consider i.i.d. observations $\{\mathbf{y}_{i}\}_{i\in\mathscr{I}},\mathbf{y}_{i}\in\mathbb{R}^n$. A VAE introduces a latent variable $Z\in\mathbb{R}^d$ with $d < n$ in a generative model $P_{Y,Z}$ to capture the underlying characteristics of the data. Specifically, the generative process draws a sample $\mathbf{z}_i$ from a prior distribution $P_Z$, typically a multivariate standard normal $P_Z=\mathcal{N}_d(\boldsymbol{0},\mathbf{I}_d)$, and then generates an observation $\mathbf{y}_i$ from the conditional sampling distribution $P_{Y\mid \mathbf{z}_i}$.
Accordingly, the density of the joint distribution factorizes as $p_{Y,Z}(\mathbf{y}_i,\mathbf{z}_i)=p_{Y\mid \mathbf{z}_i}(\mathbf{y}_i)p_{Z}(\mathbf{z}_i)$.

Within the VAE framework, the marginal likelihood of an observation, $p_Y(\mathbf{y})$, is obtained by
\begin{align*}
p_Y(\mathbf{y}_i)=\int_{\mathbb{R}^d}p_{Y\mid\mathbf{z}}(\mathbf{y}_i)p_Z(\mathbf{z}) \mathrm{d}\mathbf{z}.
\end{align*}
Therefore, evaluation of the marginal likelihood \(p_Y(\mathbf{y}_i)\) involves integrating over the latent space, making the required computations analytically intractable. This intractability extends to the posterior \(P_{Z\mid\mathbf{y}_i}\), because Bayes’ rule demands dividing by \(p_Y(\mathbf{y}_i)\). As a result, any inference or model evaluation that relies on the exact posterior or marginal likelihood becomes unfeasible.

To circumvent this problem, VAEs employ a variational distribution \(Q_{Z\mid\mathbf{y}_i}\approx P_{Z\mid\mathbf{y}_i}\), providing a tractable approximation to the true posterior.
This variational distribution is chosen as a multivariate normal distribution with independent components, whose parameters are the function output of an encoder neural network $\mathrm{enc}_{\boldsymbol{\phi}}:\mathbb{R}^n\to\mathbb{R}^d\times\mathbb{R}^d_{> 0}$ parameterized by learnable neural network parameters $\boldsymbol{\phi}$:
\begin{align*}
    Q_{Z\mid\mathbf{y}_i} = \mathcal{N}_d(\boldsymbol{\mu}_i,\mathrm{diag}({\boldsymbol{\sigma}_i^2}));\ (\boldsymbol{\mu}_i,\boldsymbol{\sigma}_i) = \mathrm{enc}_{\boldsymbol{\phi}}(\mathbf{y}_i).
\end{align*}
To enable gradient-based optimization, latent values are obtained by sampling through the reparametrization trick, with decomposition $\mathbf{z}_i = \boldsymbol{\mu}_i+\boldsymbol{\sigma}_i\odot\boldsymbol{\varepsilon}_i$, such that the gradient can be obtained from the non-random part, while $\boldsymbol{\varepsilon}_i$ is auxiliary noise drawn from a multivariate standard normal distribution $\mathcal{N}_d(\boldsymbol{0},\mathbf{I}_d)$, and $\odot$ denotes the Hadamard product between two vectors.
A decoder neural network 
$\mathrm{dec}_{\boldsymbol{\theta}}:\mathbb{R}^d\to\boldsymbol{\Psi}$, parameterized by 
$\boldsymbol{\theta}$, maps the latent values back to the data space by modeling the parameters $\boldsymbol{\psi}\in\boldsymbol{\Psi}$ of the conditional sampling distribution $P_{Y\mid\mathbf{z}_i}$. The parameters of the encoder and decoder neural networks are simultaneously optimized by estimating and minimizing the evidence lower bound (ELBO) function
\begin{align*}
\mathcal{L}_{\mathrm{VAE}}(\boldsymbol{\phi},\boldsymbol{\theta})=-\mathbb{E}_{Z\sim Q_{Z\mid\mathbf{y}_i}}\left[\log p_{Y\mid\mathbf{z}}(\mathbf{y}_i)\right] + \beta \mathrm{KL}\left[Q_{Z\mid\mathbf{y}_i}\mid\mid P_Z\right],
\end{align*}
where $\mathrm{KL}\left[\,\cdot\,\mid\mid\,\cdot\,\right]$ denotes the Kullback-Leibler (KL) divergence between the variational distribution 
$Q_{Z\mid\mathbf{y}_i}$
and the prior $P_Z$ and the parameter $\beta>0$ balances the KL-divergence with the reconstruction error term~\cite{betaVAE}. Optimization of the ELBO function is performed over mini-batches, i.e., randomly selected groups of observations, using stochastic gradient descent methods.

For patients $i\in\mathscr{I}$ that have longitudinal measurements, we denote their data as $\mathbf{Y}_i=\left(\mathbf{y}_{i,t}\right)^\top_{t\in\mathscr{T}_i} \in\mathbb{R}^{m_i\times n}$. The individual observations are collected at $m_i$ visit times $\mathscr{T}_i=\lbrace{t_{i,1},\dots, t_{i,m_i}:0= t_{i,1}<\dots< t_{i,m_i}\rbrace}$. We optimize the network weights simultaneously on a dataset containing all time points and patients, treating them as i.i.d. observations. This results in an $m_i\times d$-dimensional representation $\mathbf{Z}_i=\left(\mathbf{z}_{i,t}\right)^\top_{t\in\mathscr{T}_i} \in\mathbb{R}^{m_i\times d}$ for each patient $i\in\mathscr{I}$, as a basis for subsequent modeling by linear mixed-effects regression.

\subsection{Multivariate longitudinal mixed-effects regression}

To incorporate time structure into the latent representation, we utilize multivariate mixed-effects regression models
\cite{CHIPPERFIELD2012146, mlmm}. They represent a patient trajectory as a realization of a random variable $Z_i\in\mathbb{R}^{m_i\times d}$, which is a linear combination of fixed effects, random effects, and residual errors:
\begin{align*}
Z_i=\mathbf{X}_i\mathbf{B}+\mathbf{T}_iU_i+E_i.
\end{align*}

Fixed effects $\mathbf{B}\in\mathbb{R}^{p\times d}$ with design matrix $\mathbf{X}_i \in\mathbb{R}^{m_i\times p}$ represent the influence of $p$ patient covariates with consistent effects across all patients. Such covariates could include age, sex, or administered medications. Interaction terms with age can model a change in influence over time. 

Random effects $U_i\in\mathbb{R}^{q\times d}$ with design matrix $\mathbf{T}_i\in\mathbb{R}^{m_i\times q}$ allow for individual-specific deviations from the fixed group effects. Common choices include a random intercept and slope to model patient-specific trajectories.
In this study, we assume that the random effects follow a normal distribution $\mathrm{vec}(U_i)\sim\mathcal{N}_{qd}(\boldsymbol{0},\boldsymbol{\Phi})$ with shared covariance matrix $\boldsymbol{\Phi}$ across patients, where $\mathrm{vec}:\mathbb{R}^{n\times m}\to\mathbb{R}^{nm}$ denotes the vectorization function, which stacks the columns of a matrix into a vector.
Residual errors $E_i\in\mathbb{R}^{m_i\times d}$
account for variability not explained by fixed and random effects.
Residuals are assumed to be independent across time given the random effects and across patients, $\mathrm{Cov}\big(\mathrm{vec}(U_i),\mathrm{vec}(E_i)\big)=0$. Furthermore, we assume that they follow a normal distribution $\mathrm{vec}(E_i)\sim\mathcal{N}_{m_id}(\boldsymbol{0},\boldsymbol{\Sigma}\otimes \mathbf{I}_{m_i})$, where $\otimes$ denotes the Kronecker product between two matrices. 

The marginal distribution of the response variable $Z_i$ can be described as
\begin{align*}
\mathrm{vec}(Z_i)\sim\mathcal{N}_{m_id}\big(\mathrm{vec}(\mathbf{X}_i\mathbf{B}),\underbrace{(\mathbf{I}_d\otimes\mathbf{T}_i)\boldsymbol{\Phi}(\mathbf{I}_d\otimes\mathbf{T}_i)^\top+\boldsymbol{\Sigma}\otimes \mathbf{I}_{m_i}}_{=:\mathbf{V}_i^{-1}}\big)
\end{align*}
Given observed patient trajectories $\mathbf{Z}_i$ of the random variable $Z_i$, the variance parameters $\boldsymbol{\Phi}$ and $\boldsymbol{\Sigma}$ are estimated using maximum likelihood (ML) or restricted maximum likelihood (REML) estimation~\cite{sandsplusmixedmodels}, where the log-likelihoods can be calculated as
\begin{align*}
\mathcal{L}_{\mathrm{ML}}\big(\mathbf{B},\boldsymbol{\Phi}, \boldsymbol{\Sigma}\big) =& \frac{1}{2} \sum_{i\in\mathscr{I}} \Big( \log\det(\mathbf{V}_i) - \mathrm{vec}(\mathbf{Z}_i -\mathbf{X}_i\mathbf{B})^\top \mathbf{V}_i \mathrm{vec}(\mathbf{Z}_i -\mathbf{X}_i\mathbf{B})\\ &- m_i d \log(2\pi) \Big)\\ 
\mathcal{L}_{\mathrm{REML}}\big(\boldsymbol{\Phi}, \boldsymbol{\Sigma}\big) =& \frac{1}{2} \sum_{i\in\mathscr{I}} \Big(\log\det(\mathbf{V}_i) - \log\det\big((\mathbf{I}_d\otimes\mathbf{X}_i)^\top \mathbf{V}_i (\mathbf{I}_d\otimes\mathbf{X}_i)\big) \\&-  \mathrm{vec}(\mathbf{Z}_i -\mathbf{X}_i\widehat{\mathbf{B}})^\top  \mathbf{V}_i \mathrm{vec}(\mathbf{Z}_i -\mathbf{X}_i\widehat{\mathbf{B}}) - (m_i-p) d \log(2\pi) \Big). \end{align*} 
Optimization algorithms like Newton-Raphson or quasi-Newton methods can be used to maximize the likelihood functions~\cite{mixexdmodel_optim}.
With an estimate of the covariance components and observed response $\mathbf{Z}_i$,
the realized best linear unbiased estimator (BLUE) $\widehat{\mathbf{B}}$ of the fixed effects, is given by \begin{align*} 
\mathrm{vec}(\widehat{\mathbf{B}}) = \left( \sum_{i\in\mathscr{I}} (\mathbf{I}_d\otimes\mathbf{X}_i)^\top \widehat{\mathbf{V}}_i (\mathbf{I}_d\otimes\mathbf{X}_i) \right)^{-1} \left( \sum_{i\in\mathscr{I}} (\mathbf{I}_d\otimes\mathbf{X}_i)^\top \widehat{\mathbf{V}}_i \mathrm{vec}(\mathbf{Z}_i) \right). 
\end{align*}
The best linear unbiased predictor (BLUP) of the random effects $U_i$ can be obtained through
\begin{align*} 
\mathrm{vec}(\widehat{\mathbf{U}}_i) =
\widehat{\boldsymbol{\Phi}} (\mathbf{I}_d\otimes\mathbf{T}_i)^\top \widehat{\mathbf{V}}_i \,\mathrm{vec}(\mathbf{Z}_i -\mathbf{X}_i\widehat{\mathbf{B}}).
\end{align*}

\subsection{Joining multiple measurement instruments in a latent representation}
\label{multiple_VAEs}

\begin{figure}[!ht]
    \centering
    \includegraphics[width=\textwidth]{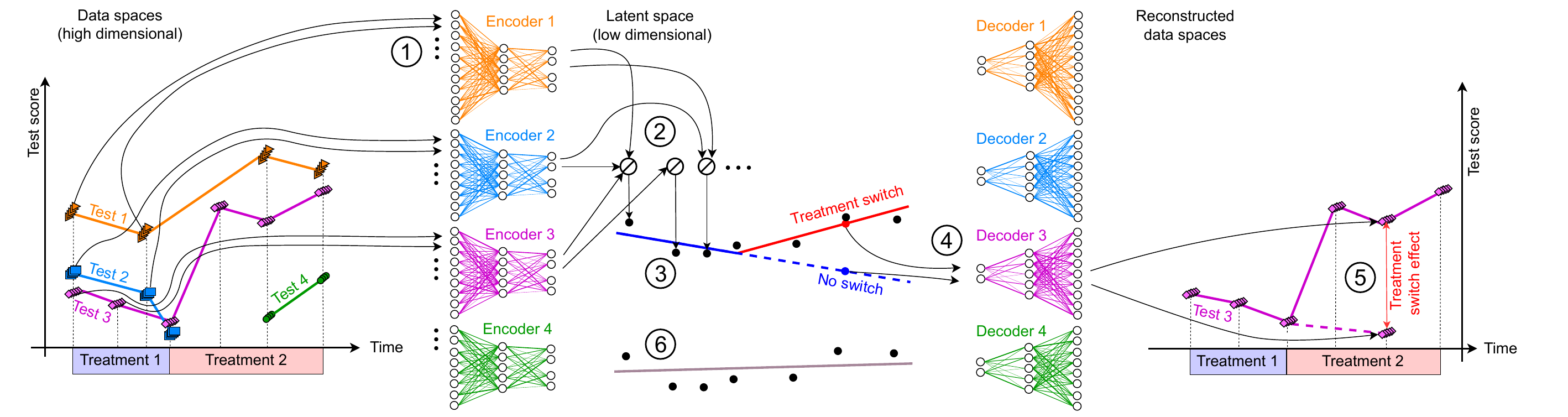}
    \caption{Schematic illustration of the proposed approach for multiple measurement instruments: 
    1) The items of each measurement instrument are encoded by a dedicated encoder network and corresponding latent values are drawn. 
    2) The latent values are averaged across instruments at each time step, to obtain a joint latent trajectory.
    3) The averaged latent values serve as outcome variable for a multivariate mixed-effects regression, which provides BLUE and BLUP estimators.
    4) Predictions from the mixed-effects model serve as input to a dedicated decoder network for each measurement instrument, for reconstruction at the original item level. 
    5) Treatment switch effects can be quantified per item.
    6) A likelihood-ratio test provides statistical inference.}
    \label{fig:approach}
\end{figure}

For incorporating data from multiple measurement instruments $l\in\mathscr{L}\subset\mathbb{N}$, where each instrument comprises $n_l$ test items, we use one encoder and decoder network per instrument, embedding data into a joint latent representation. Here observations are linked by multivariate mixed-effects regression, as schematically shown in Figure \ref{fig:approach}. 
We assume a patient $i\in\mathscr{I}$ is assessed using a subset of 
different measurement instruments at each time point. Denote by $\mathscr{T}_{i,l}\subseteq\mathscr{T}_{i}$ the observations in which measurement instrument $l\in\mathscr{L}$ was utilized.
At the data level we obtain one trajectory per patient and observed instrument,
\begin{align*}
\mathbf{Y}_{i,l}=\left(\mathbf{y}_{i,t,l}\right)^\top_{t\in{\mathscr{T}_{i,l}}} \in\mathbb{R}^{m_{i,l}\times n_l}, \quad l\in\mathscr{L},
\end{align*}
where $m_{i,l}$ denotes the overall number of observations for patient $i$ with measurement instrument $l$. 
Measurement instrument specific encoder networks $\mathrm{enc}_{\boldsymbol{\phi_l}}$ are used to obtain the parameters of the variational distributions $Z_l\sim Q_{Z_l\mid\mathbf{y}_{i,t,l}}$ from which we draw $d$-dimensional realizations 
\begin{align*}
\left(\mathbf{z}_{i,t,l}\right)^\top_{t\in{\mathscr{T}_{i,l}}} \in\mathbb{R}^{m_{i,l}\times d}, \quad l\in\mathscr{L}.
\end{align*}

Denote with $\mathscr{L}_{i,t}\subseteq\mathscr{L}$ the index set of all measurement instruments used for patient $i$ at observation time $t$. To obtain a unified latent trajectory for a patient, we average latent variables over all available values at the respective time points, yielding
\begin{align*}
\mathbf{Z}_{i} = (\mathbf{z}_{i,t})_{t\in \mathscr{T}_i}^\top = \left(\sum_{l\in \mathscr{L}_{i,t}} \frac{\mathbf{z}_{i,t,l}}{\lvert\mathscr{L}_{i,t}\rvert}\right)_{t\in \mathscr{T}_i}^\top\in\mathbb{R}^{m_i\times d}.
\end{align*}

We then model the longitudinal structure in the latent space using a multivariate mixed-effects model of dimension $d$. 
The latent mixed-effects model in particular can incorporate effects of treatment switches. Let $t_{\mathrm{switch}_i}$ denote the time of a switch from one treatment to another. We collect the time differences relative to the switch 
\begin{align*}
\Delta\mathbf{t}_i = \big(t_{i,1}-t_{\mathrm{switch}_i}, t_{i,2}-t_{\mathrm{switch}_i},\ldots, t_{i,m_i}-t_{\mathrm{switch}_i}\big)^\top.
\end{align*}
To capture group-level trends following a treatment switch, the time since switch $\max(\boldsymbol{0},\Delta\mathbf{t}_i)$ is included as a fixed effect. To model individual deviations from group trends, there are random effects included for the time since $\max(\boldsymbol{0},\Delta\mathbf{t}_i)$ and the time up to a treatment switch $\min(\boldsymbol{0},\Delta\mathbf{t}_i)$. When multiple treatments are involved their influence can be separated into treatment specific covariates. Interaction terms with other covariates (e.g., age) can also be included in the mixed-regression model.
All decoder networks receive as input the prediction of the mixed model
\begin{align*}
\widehat{\mathbf{Z}}_i=\mathbf{X}_i\widehat{\mathbf{B}}+\mathbf{T}_i\widehat{\mathbf{U}}_i
\end{align*}
where $\widehat{\mathbf{B}}$ is the BLUE and $\widehat{\mathbf{U}}_i$ is the BLUP, given the latent outcome $\mathbf{Z}_i$.

To estimate the parameters of the encoders and decoders as well as the latent mixed-effects model, we use an iterative approach: To obtain the encoder and decoder parameters, $(\phi_l,\theta_l)$ for $l\in\mathscr{L}$, we freeze the mixed-effect model parameters and minimize as a combination of per-measurement instrument ELBO loss functions and two alignment terms:
\begin{align*}
\mathcal{L}_{\mathrm{MMVAE}}\big((\boldsymbol{\phi}_l, \boldsymbol{\theta}_l)_{l\in\mathscr{L}}\big)=\\\frac{1}{|\mathscr{I}|\sum_{i\in\mathscr{I}}|\mathscr{T}_{i}|}\bigg(\gamma\sum_{i\in\mathscr{I}}\sum_{t\in \mathscr{T}_{i}} \|\widehat{\mathbf{z}}_{i,t}-\mathbf{z}_{i,t}\|^2_2-\eta\mathcal{L}_{\mathrm{ML}}(\mathbf{B},\boldsymbol{\Phi},\boldsymbol{\Sigma})\bigg)+\\
\frac{1}{|\mathscr{I}|\sum_{i\in\mathscr{I}}\sum_{l\in\mathscr{L}}|\mathscr{T}_{i,l}|}\sum_{i\in\mathscr{I}}\sum_{l\in\mathscr{L}}\sum_{t\in \mathscr{T}_{i,l}}\bigg(-\mathbb{E}_{Z_l\sim Q_{Z_l|\mathbf{y}_l}}\left[\log p_{Y_l|\widehat{\mathbf{z}}_{i,t}}\left(\mathbf{y}_{i,t,l}\right)\right]+\\\beta \mathrm{KL}\left[Q_{Z_l|\mathbf{y}_{i,t,l}}||P_Z\right]\bigg)
\end{align*}

The first alignment term, weighted by $\eta\ge0$ encourages agreement with the training criterion (ML/REML) of the frozen mixed model, while the second alignment term weighted by $\gamma\ge0$, encourages agreement between the encoder output and the latent mixed model prediction.

Because the joint latent, serving as response to the mixed model, is the average of measurement specific latent variables $\mathbf{z}_{i,t}=|\mathscr{L}_{i,t}|^{-1}\sum_{l\in\mathscr{L}_{i,t}}\mathbf{z}_{i,t,l}$, the encoders are implicitly driven to place their outputs on a common scale. Instrument-specific shifts or a rescaling inflates the disagreement penalty $\|\widehat{\mathbf{z}}_{i,t}-\mathbf{z}_{i,t}\|_2^2$, as well as reduce the mixed-effects likelihood, and as mixed model predictions are passed to the decoder ultimately the reconstruction loss and are therefore corrected during training.

After fitting the encoder and decoder parameters for given mixed-effect model parameters for a given amount of iterations with a gradient descent method like Adam~\cite{kingma2017adam}, a latent representation for each trajectory $\mathbf{Z}_i,i\in\mathscr{I}$ is drawn. Afterwards, the mixed-effects model is fitted using these latent representations as response, with ML or REML estimation.
Such alternating updates of the VAE and mixed-effects model parameters are performed until the loss function $\mathcal{L}_{\mathrm{MMVAE}}$ no longer steeply decreases and decreases saturate.

With the final parameter estimates, the impact of treatment switches can be quantified on the item level by decoding from the latent trajectory with and without treatment switch. These trajectories can be obtained by calculating the BLUE and BLUP, modifying the design matrices $\mathbf{X}_i,\mathbf{T}_i$ by setting the covariates and interaction terms relating to time elapsed after treatment switches to zero and extending the covariates relating to time before treatment switches. The latent prediction of both models can be mapped back to the data level using the decoder neural networks and analyzed at the item level for differences.

\subsection{Model selection}

Statistical testing based on the mixed-effects model can be useful for further assessing the effect of treatment switches, as well as for potential model selection, e.g., to decide which covariates to include for adjustment. One would normally assess the significance of a covariate by a likelihood-ratio test~\cite{wilkes} with test-statistic $\Lambda \;=\; 2(\mathcal{L}_{\mathrm{ML}\text{ full}}-\mathcal{L}_{\mathrm{ML}\text{ red}})$. Here, $\mathcal{L}_{\mathrm{ML}\text{ full}}$ denotes the likelihood of a model containing the covariate of interest and
$\mathcal{L}_{\mathrm{ML}\text{ red}}$ corresponds to the likelihood of a nested model without it. Under Wilks’ theorem and standard regularity $\Lambda$ is asymptotically $\chi^2_{rd}$ for a fixed effects block, with $r$ being the number of tested covariates and $d$ the dimension of the latent space.  

The classical statistical model selection approach cannot be applied here as the final latent responses $\mathbf Z_i$ depend on encoder parameters $\boldsymbol\phi_l,l\in\mathscr{L}$ that were jointly optimized with the mixed-effects coefficients. The same data is therefore used first to pick a representation and then to test a hypothesis, which can induce a post-selection bias~\cite{taylor2015statistical, lee2016exact}. In this setting the $\chi^{2}_{rd}$ reference distribution can be invalid. To correct for the intractable selection event (“the VAEs picked a particular latent representation”), we combine knockoff negative-control covariates~\cite{barber2015controlling, nguyen2020ako} with bootstrap~\cite{efron1979bootstrap,taylor2015statistical}. The resulting procedure allows for inference for the joint VAE–mixed-effects estimator.

Let $\mathbf{w}_{i}\in\mathbb{R}^{k}$ denote the realization of a random
variable $W_i$, drawn independently of
$\{(\mathbf{Z}_{i},\mathbf{X}_{i},\mathbf{T}_{i})\}_{i\in\mathscr{I}}$ with
$\mathbb{E}[W_{i}]=\boldsymbol{0},\;
\mathrm{Cov}(W_{i})=\mathbf{I}_k$.
Let $\mathbf{W}_{i} = (\mathbf{w}_{i,1}\cdots\mathbf{w}_{i,m_i})^\top$ be concatenated variables that are either equal for each visit for patient-level, or re-drawn to represent visit-level knockoff variables. 
We augment the fixed effect design matrix by
$\widetilde{\mathbf{X}}_{i}=[\mathbf{X}_{i}\;\mid\mathbf{W}_{i}\;]$ and test
\begin{align*}
\label{eq:knockoffH0}
H_{0}:\;\boldsymbol{\beta}_{\!W}= \boldsymbol{0}
\quad\text{vs.}\quad
H_{1}:\;\boldsymbol{\beta}_{\!W}\neq\boldsymbol{0},
\end{align*}
where $\boldsymbol{\beta}_{\!W}\in\mathbb{R}^{K\times d}$ denotes the
coefficient block of~$\mathbf{W}_{i}$.  Because $\mathbf{W}_{i}
\perp\nobreak\!\!\!\perp\nobreak
\mathbf{Z}_{i}\,\mid\,\mathbf{X}_{i},\mathbf{T}_{i}$ by construction, $H_{0}$ is
true with probability 1 and $\mathbf{W}_{i}$ acts as a negative-control predictor in the sense of the work by~\cite{barber2015controlling}, yielding a reference statistic that captures what one should observe under the null hypothesis. 

To empirically recreate the intractable selection event, we construct an empirical null distribution for the likelihood ratio test $F_{\Lambda}^{0}$ via bootstrap. For bootstrap samples $b\in\mathscr{B}$, we randomly reinitialize the VAE parameters and regenerate knockoff variables $\mathbf{W}_{i,b}$. Afterwards we re-optimize for $\{(\boldsymbol{\phi}_{l,b},\boldsymbol{\theta}_{l,b})\}_{l\in\mathscr{L}}$ to obtain latent responses $\mathbf Z_{i,b}, b\in\mathscr{B}$, with which we calculate the likelihood ratio statistic $\Lambda_{b}$ from the fitted full and reduced models.
The empirical cumulative distribution function\ 
\begin{align*}
\widehat{F}_{\Lambda}^{0,\mathscr{B}}(x)=
\frac1{|\mathscr{B}|}\sum_{b\in\mathscr{B}}\mathbbm{1}\{\Lambda_{b}\le x\}
\end{align*}
converges in probability to~$F_{\Lambda}^{0}$ as
$|\mathscr{B}|\to\infty$, while resampling the knockoff together with the data (via different network weights) preserves the exchangeability property that underlies false discovery rate control in the original knockoff filter~\cite{nguyen2020ako}. Consequently, the $p$-value
\begin{align*}
\hat{p}
=\frac{1+\sum_{b\in\mathscr{B}} \mathbbm{1}\{\Lambda_{b}\ge\Lambda_{\mathrm{obs}}\}}
      {|{\mathscr{B}}|+1}
\end{align*}
is an unbiased estimate of $\mathbb{P}_{0}(\Lambda\ge\Lambda_{\mathrm{obs}})$.

\section{Results}
\label{results}

We use the proposed approach on data from the SMArtCARE registry~\cite{SMArtCARE}, which tracks the disease course of patients with spinal muscular atrophy (SMA). 
SMA is a genetic disorder characterized by progressively declining motor function with symptom onset at birth or within early childhood. The severity of symptom progression is influenced by patient characteristics like genetic markers (e.g., count of the \textit{SMN2} gene) or age at symptom onset and varies significantly across individuals~\cite{SMA_characteristics}. 
Recently, there were advances in treating SMA~\cite{SMA_medication}, leading patients to switch to emerging treatments when they became available to them. Treatment assignment within the SMArtCARE registry is not randomized, yet mainly depends on the availability of a treatment in a medical center. To monitor the impact of treatment switches on disease development, longitudinal motor function assessments are conducted during semi-regular visits. Here, movement ability is assessed using five specialized measurement instruments, the CHOP-INTEND~\cite{CHOP}, HINE-2~\cite{HINE}, HFMSE~\cite{HFMSE}, RULM~\cite{RULM} and ALSFRS-R~\cite{ALSFRS} motor function tests.  (See  \hyperref[tab:measurement]{Table \ref*{tab:measurement}} for details). 
In practice, infants and weak non-sitters are evaluated with CHOP-INTEND and motor milestone attainment in early childhood is captured with HINE-2. When patients age, assessments transition to HFMSE to assess gross motor ability and RULM for upper-limb performance, with ALSFRS-R complementing motor scales in adult cohorts.  
Instrument changes often occur to use stage-appropriate instruments and to avoid flooring and ceiling effects.

Our approach integrates these five different motor function measurement instruments for a cohort of $522$ patients, with a median of $17$ individual observations per patient and mean of $2.1$ out of $5$ measurement instruments used per visit. 
The latent mixed-effects regression model incorporates as covariates the time elapsed since treatment switch, age, a genetic marker, specifically the \textit{SMN2} count, age at symptom onset, an indicator whether patients are currently asymptomatic for example because they were diagnosed via newborn screening, the current ventilation status, the current scoliosis status, sex of the patient and if a family member is affected by SMA. We also include interactions with age as fixed effects. Additionally, three random-effect terms were used to allow for individual deviations from group-level trends as random effects: a random intercept, and random slopes for the trajectory before and after the treatment switch. 

\subsection{Quantifying the impact of treatment switches}

A key question in the context of SMA treatment is the impact of treatment switches on the individual and the population level. To assess this effect in the light of different measurement approaches, we use the proposed approach to compare two sets of predicted disease trajectories: one incorporating the treatment switch and a scenario in which the switch did not occur. We decoded the mixed-effects model predictions from both scenarios for one year after the treatment change and compared their reconstructed motor function test item scores. The difference in total sum scores then served as our estimate of the treatment switch effect.

As seen in Table \ref{tab:data_level_effect}, there is a positive predicted improvement between $1.8\%$ and $5.4\%$ of the maximal score across the measurement instruments. Results are not deterministic as the encoder and decoder model distributions introduce additional stochasticity in the sampling and fitting process. However, assessed over ten random initializations there was no occurrence in which the aggregated treatment effect switched from positive to negative in one of the assessed instruments. 

To assess sensitivity, we simulated an additional artificial treatment switch by adding a sum score of $+2$ points, distributed to randomly selected test items for each year after the switch (see chapter \ref{params} for details). However, this perturbation ignores ceiling effects as scores cannot exceed the sum of the maximal individual item scores, within-observation structure, and cross-instrument alignment. We therefore expect the approach to detect the switch but understate its magnitude.
Consistent with this expectation, the approach recovered an average treatment switch effect between $0.78$ and $1.97$ on the measurement instruments, indicating correct directionality and no overestimation of the effect size. 

\begin{figure}[!ht]
    \centering
        \includegraphics[width=\linewidth]{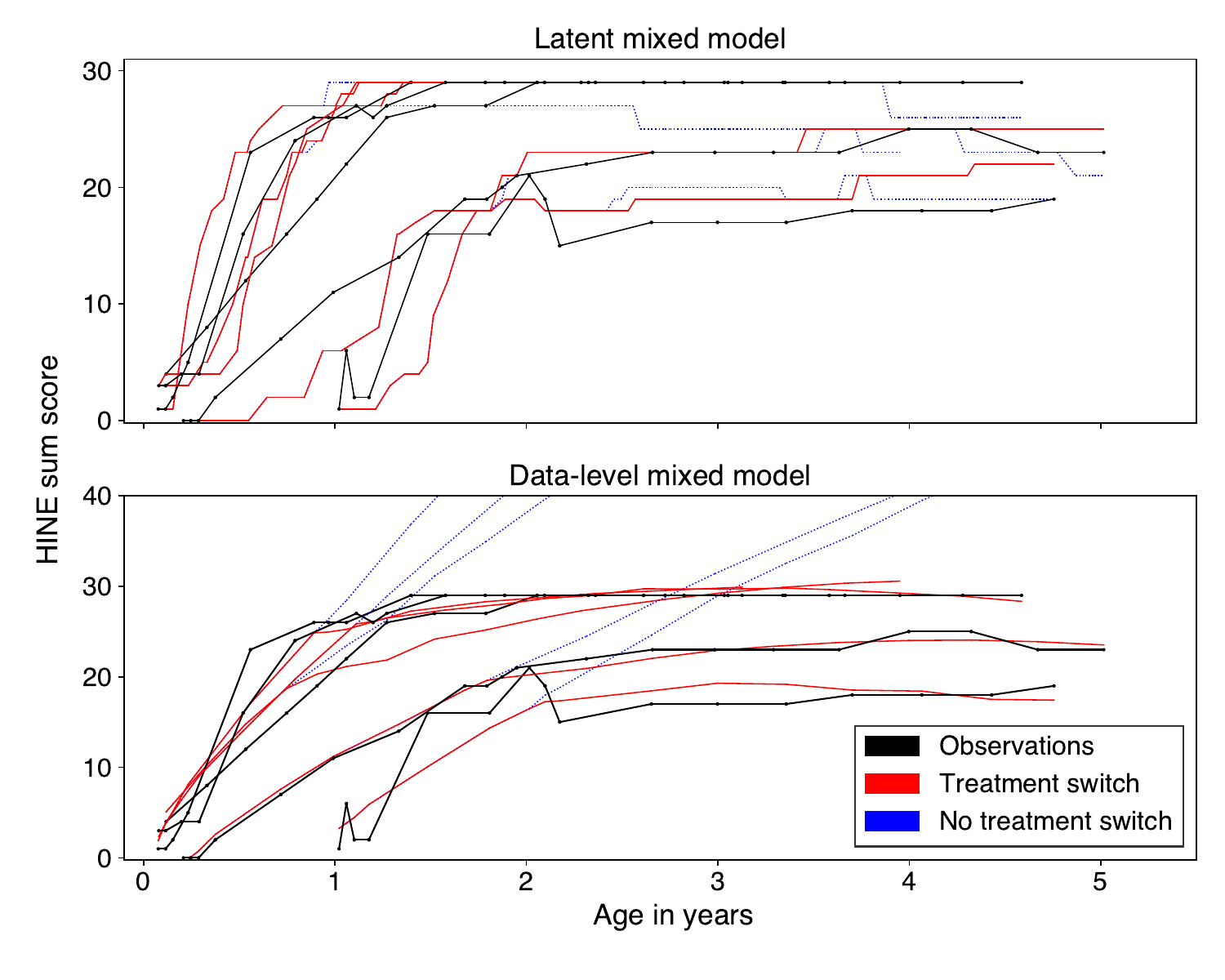}
    \caption{Comparison of total HINE-2 sum scores for five patient trajectories. Observed data trajectories are displayed in black, the respective mixed model prediction with treatment switch in red, and mixed model predictions for a hypothetical trajectory without treatment switch in blue. The reconstructed trajectories of the latent mixed model are displayed in the upper subplot and for a standard data-level mixed model in the lower subplot.}
    \label{fig:ceiling}
\end{figure}

We also found that our approach reduces ceiling effects compared with a data-level mixed model without additional parameters to account for ceiling or floor effects (see Figure \ref{fig:ceiling}). Because the linear latent mixed-model predictions are passed through a decoder neural network which models the data distribution, samples are restricted to consist of plausible reconstructions. Non-linearity is further introduced by neural network transformations.

\subsection{Latent model selection}

\begin{figure}[!ht]
    \centering
    \begin{subfigure}{0.49\textwidth}
        \centering
        \includegraphics[width=\linewidth]{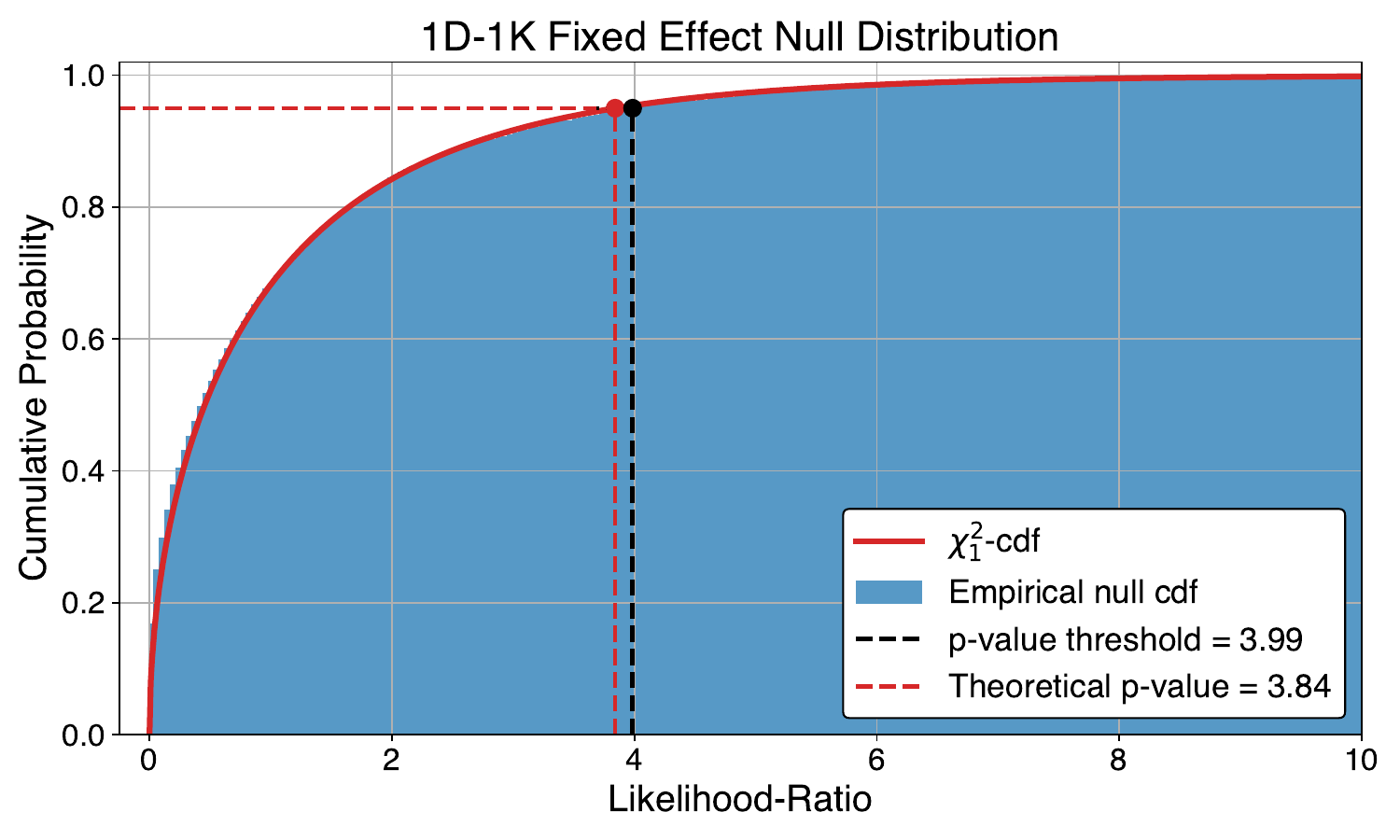}
    \end{subfigure}
    \begin{subfigure}{0.49\textwidth}
        \centering
        \includegraphics[width=\linewidth]{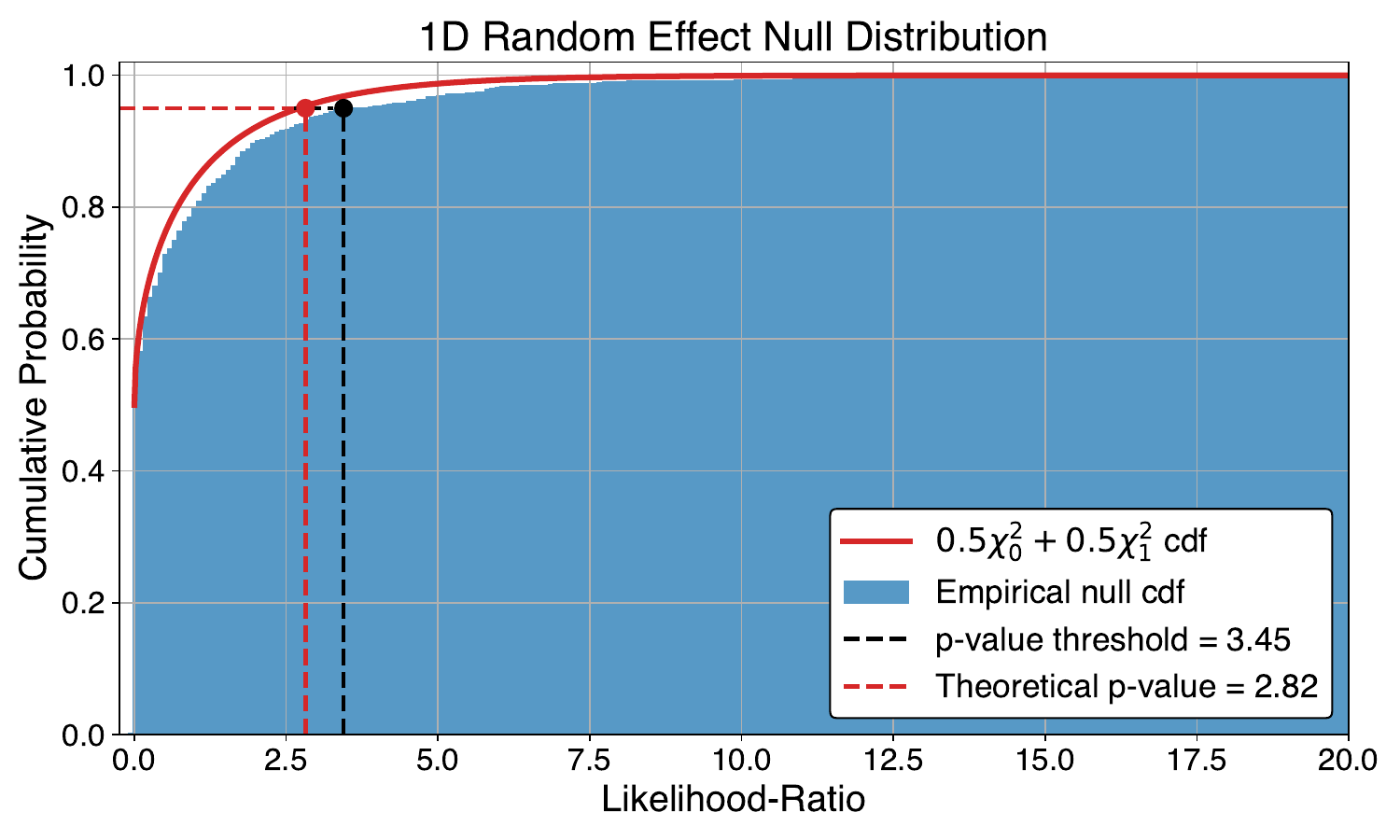}
    \end{subfigure}
    \begin{subfigure}{0.49\textwidth}
        \centering
        \includegraphics[width=\linewidth]{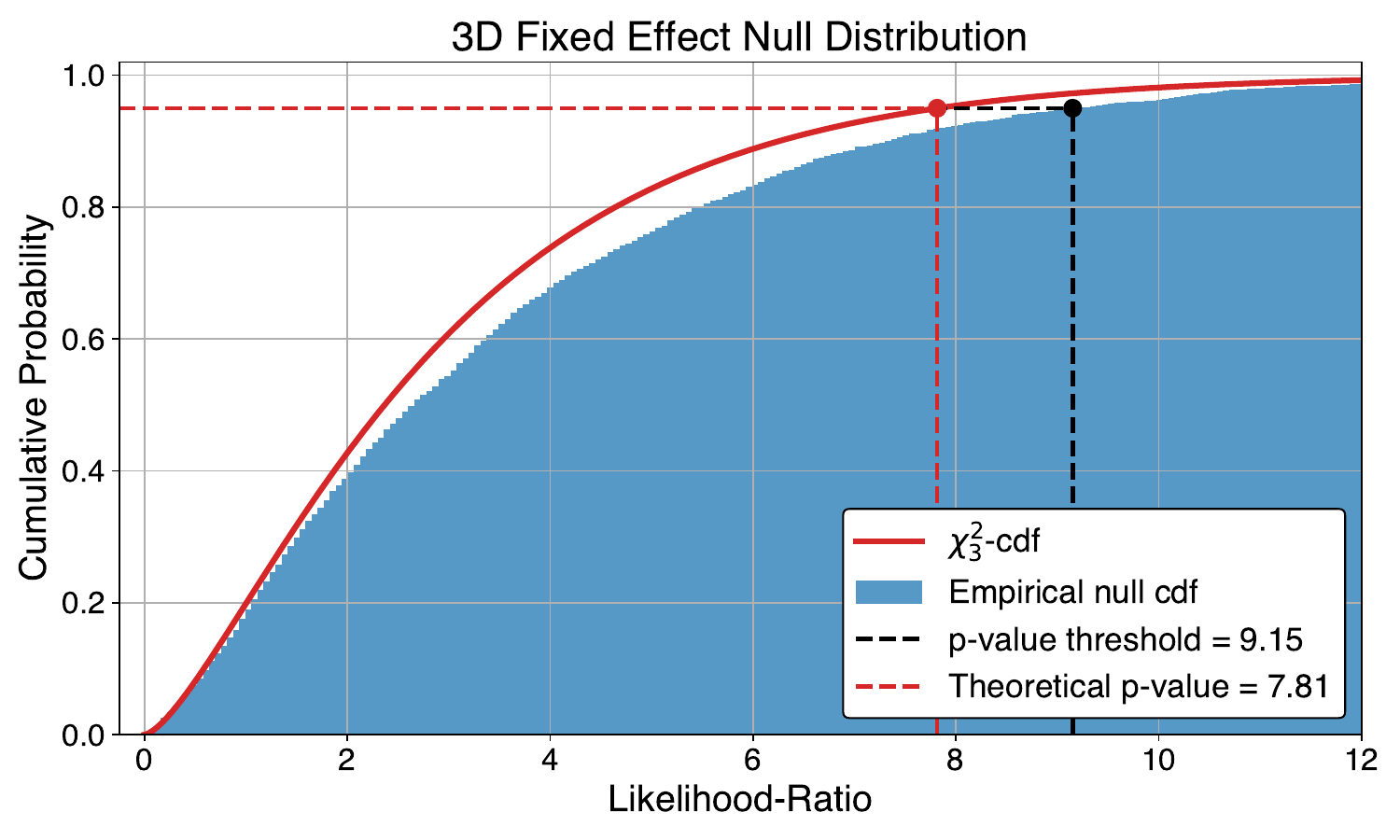}
    \end{subfigure}
    \begin{subfigure}{0.49\textwidth}
        \centering
        \includegraphics[width=\linewidth]{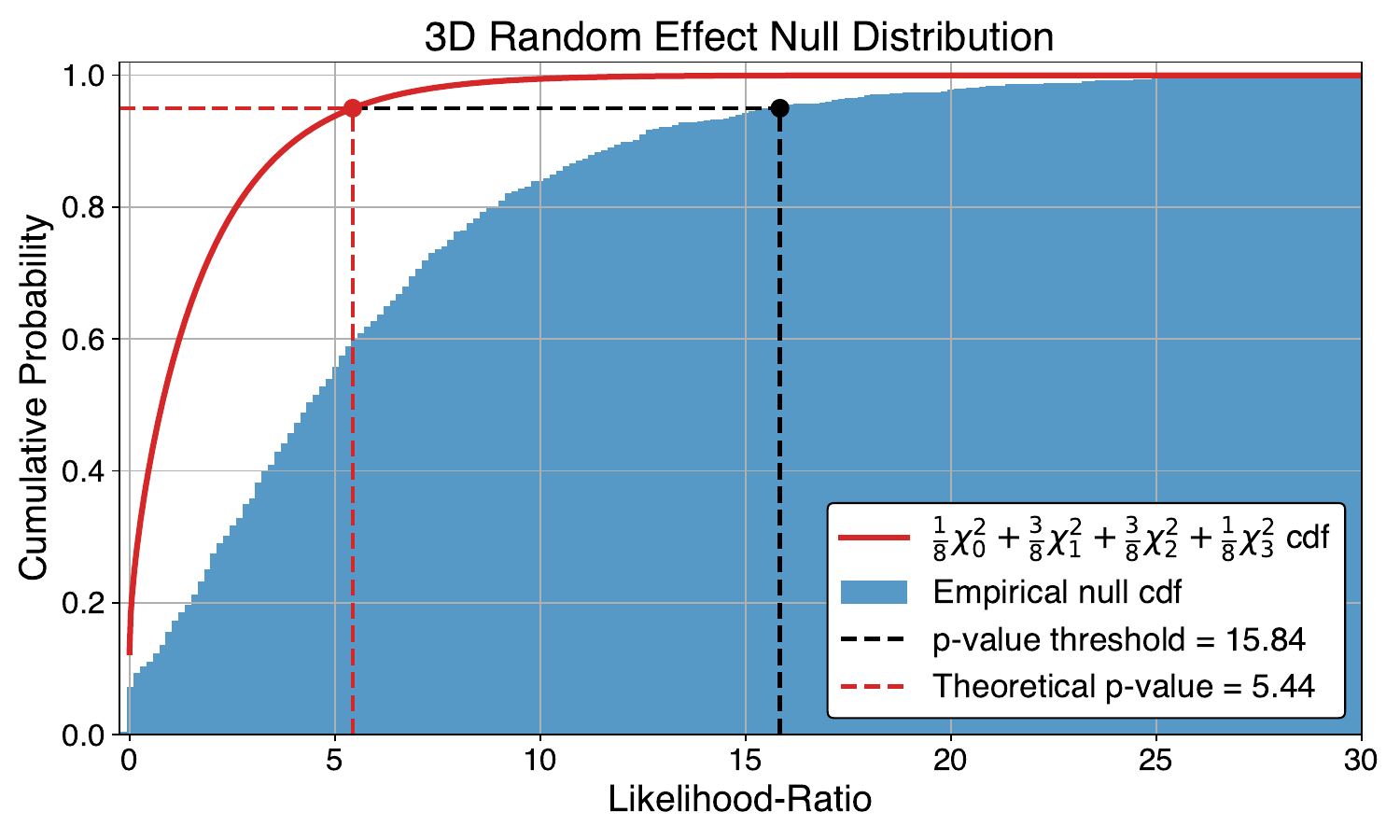}
    \end{subfigure}
    \caption{Empirical cumulative distribution functions of the null distributions of the likelihood ratio test statistic from artificially added knockoff variables for fixed (left column) and random effects (right column) in the mixed-effects regression in latent representations of dimension $d=1$ (top row) and $d=3$ (bottom row). The red line represents the theoretical chi-squared distribution that does not take into account the interdependent VAE and mixed model training procedure.}
    \label{fig:empiricalcdf}
\end{figure}

We evaluated the statistical significance of effects in the mixed-effects model, such as of treatment switches, with likelihood ratio (LR) tests with critical values obtained from the proposed knockoff variable bootstrap procedure. The empirical distribution of the test statistic was constructed with $|\mathscr{B}|=1000$ bootstrap samples.

First, we compared the empirical null distributions with the theoretical $\chi^2$-distributions that ignore the influence of the VAEs (see Figure \ref{fig:empiricalcdf}). For a one-dimensional latent representation, the empirical knockoff distribution follows the theoretical distribution closely. However, for the three-dimensional representation that we want to use, there is a considerable difference, where the theoretical distribution would lead to an anti-conservative test. This is seen for fixed effects as well as for random effects, where deviations are even larger. 

Subsequently, we conducted LR tests for the $11$ fixed effects covariates in the mixed-effects model. We indicate with $rd$ the degrees of freedom the theoretical $\chi^2$-test statistic would follow, consisting of the amount of differences in included fixed effect ($r$) multiplied with the latent dimension ($d$). The effects for ventilation status ($\Lambda_{\mathrm{obs}}=182.8, \mathrm{SD}=28.0, rd=6$), and for treatment switches ($\Lambda_{\mathrm{obs}}=335.2, \mathrm{SD}=59.8, rd=12$), and disease onset ($\Lambda_{\mathrm{obs}}=237.6, \mathrm{SD}=29.7, rd=12$) and surgery for scoliosis ($\Lambda_{\mathrm{obs}}=85.3, \mathrm{SD}=14.36, rd=6$) showed high LR-values, above critical values from the knockoff null distribution. Other covariates, such as sex ($\Lambda_{\mathrm{obs}}=17.3, \mathrm{SD}=5.4, rd=6$) and family history ($\Lambda_{\mathrm{obs}}=11.2, \mathrm{SD}=4.3, rd=6$), did not display significant effects.

As a further check, we generated data using the generative nature of the VAEs based on the fitted latent trajectories, but without the treatment switch effect, and refitted the mixed-effects model to this newly generated dataset. The likelihood ratio for including the treatment switch parameters dropped to $18.2$, which no longer exceeded the empirical significance threshold ($23.6$ for $rd=12$, meaning $d=3$ latent dimensions with $r=4$ differences in covariates, consisting of switches to two SMA medications and their interactions with age).

\subsection{Comparison with naive meta-analysis}

As a comparison to the proposed approach, we fit separate linear mixed-effects models for each motor instrument at the original data level, using the same fixed and random effects as in the latent model and taking instrument-specific sum scores as outcomes. Additionally, we added an intercept to the fixed effects as test sum scores are not centered around zero compared to latent variables in our latent mixed model. Because many instruments are not observed with sufficient pre- and post-switch visits for each patient, the effective sample is smaller than for the latent approach. Whereas the latent model could use a sample size of $522$ patients, the data-level models were restricted to $308$ patients for the RULM measurement instrument, $318$ for HINE-2, $239$ for HFMSE, and to $158$ for CHOP-INTEND. For the ALSFRS-R test, too few patients ($61$) were available to fit a mixed model using the same set of covariates.

To pool treatment switch effects across instruments while accounting for overlapping patients, we performed a correlated-effects fixed-effect meta-analysis via generalized least squares, using a patient-level bootstrap to estimate the full cross-instrument covariance and a bootstrap calibration for inference. For the combined treatment effects, the global likelihood-ratio statistic was $214.06, rd=3$ with a bootstrap-calibrated p-value of $p<10^{-3}$ (no exceedance out of $1000$ null replicates). Therefore, the meta-analysis provides strong evidence for a non-zero treatment switch effect, consistent in magnitude with the latent mixed-effects regression model.

When randomly sampled to a third of the original sample size, the latent approach could still reliably detect the treatment switch ($\Lambda_{\mathrm{obs}}=102.5$), while the sample size was now often too small to get a stable fit for the data-level approach, particularly for the CHOP-INTEND measurement instrument.

\section{Discussion}
\label{discussion}

In clinical studies of rare diseases, it is essential to account for changing measurement strategies, e.g., to assess treatment effects. As patient trajectories get longer and more observations become available, it becomes more likely that different measurement instruments are employed to monitor outcomes. This could be due to a changing state of the art or that measurement instruments are conditioned on patient characteristics. To exploit all historical data collected under different settings, novel methods are necessary. Here, we proposed an approach that combines VAEs with a latent longitudinal mixed-effects regression model to capture disease progression and treatment switches in SMA data collected from multiple measurement instruments. By using encoder neural networks to map items from these instruments onto a unified latent representation, we handle systematic changes in measurement instruments during the observation period. To enable statistical inference, we introduced a knockoff variable bootstrap testing approach which allows us to correct for potential biases arising during model fitting. 

This provides unique capabilities, which we demonstrated when assessing effects of treatment switches in patients with SMA. By mapping to a latent embedding via artificial neural networks, we utilize the item-level details of observations. The learnable network mappings also handles ceiling effects and aligns measurement instruments of different scales. This allowed us to integrate observations of different measurement instruments, e.g., tailored to different age ranges, to leverage a larger effective sample size compared to traditional methods that analyze each instrument separately. In other settings, power gains are particularly relevant for evaluating treatments in rare disease registries, where data is scarce. Potential bias in subsequent statistical inference, incurred by the flexibility of neural networks, could be mitigated by the proposed knockoff variable approach. At the same time, this approach allows for the detection of treatment switch effects in the SMA application, which might not have been identified otherwise.

Still there are several aspects that need to be carefully considered when utilizing the proposed approach. Although the projection to a low-dimensional space reduces the number of required parameters, the latent dimensionality is still restricted to avoid over-parametrization of the latent model, which can limit the modeled complexity of the underlying dynamics. On the other hand, choosing the latent dimension too large can increase the effects of overfitting, i.e., a larger deviation of the skewed test statistic from the theoretical distribution. Thus, the number of latent dimensions needs to be carefully chosen, ideally depending on the dataset and based on domain knowledge. One potential approach could be to estimate the intrinsic dimension (ID)~\cite{ID_estimation} in a higher-dimensional latent space and then reduce the latent dimensionality to match the estimated ID. Further, specification of the mixed-effects model for the latent representation should incorporate biomedical knowledge, supported by the proposed statistical testing approach for model selection where necessary. This might even be an opportunity to gain further insight into the joint structure underlying several measurement instruments.

As datasets grow in complexity but not necessarily in size, which is typical for many rare diseases, there will be an increasing need for approaches that integrate deep learning methods and classical statistical theory. Such hybrid approaches can facilitate modeling on small but heterogeneous, high-dimensional and multi-modal datasets and provide answers for research questions that were previously challenging to address given the limited and complex data. As the explicit incorporation of statistical inference within deep learning frameworks is still uncommon, we therefore see this as a promising avenue of further research.

\printbibliography

\section{Funding}

Funded by the Deutsche Forschungsgemeinschaft (DFG, German Research Foundation) – Project-ID 499552394 – SFB 1597 Small Data.

\section{Data preprocessing and Hyperparameters}
\label{params}

In collaboration with our clinical partners from the SMArtCARE registry, we restricted our analysis to patients who underwent a treatment switch. To define this cohort, we filtered these patients based on the following criteria. First, patients who received less than six months of treatment with a medication before or after the switch were excluded. For patients who switched medication more than twice, we included their data only up to the second treatment switch. Further, we required that patients had motor-function evaluations at a minimum of four visits, including at least two visits before and after the switch. Lastly, we excluded switches to medications with fewer than $10$ patients, which left the dataset with switches to two different SMA medications.

This resulted in $522$ patients with an average of $17.3$ hospital visits and a mean of $2.1$ out of $5$ tests used to evaluate motor function at each visit. The median period under observation period was $5.7$ years, while the median patient age was $6.7$ years.

We binarized each test item using thermometer encoding, since physicians grade motor function on ordinal categorical scales comprising between two and six levels per test item. We also included an indicator variable for each motor function item that was set to one if a test item could not be performed by a patient. To capture the data distribution, we chose an ordinal logistic model as decoder distribution~\cite{NAZABAL2020107501}.

Missing values within an individual assessment were imputed by zero-filling, reflecting the judgment of our clinical partners that such items were typically skipped due to patient exhaustion. However, observations with more than a quarter of missing values were excluded from analysis, as some child patients occasionally refuse to perform a larger set of tasks due to bad mood despite being physically able to perform them.

Some covariates were split up into multiple fixed effects; these included the \textit{SMN2} count, which was one-hot encoded into two categories (\textit{SMN2}$\leq 2$ and \textit{SMN2}$\geq 3$). Age at symptom onset was divided into a continuous variable representing the age at symptom onset in years for symptomatic patients and an indicator variable showing whether a patient was still presymptomatic. Years since treatment switch was divided into two separate variables each indicating the years since switching to medication A and B. We standardized the covariates except time since medication switch and included interaction terms with age for all fixed effects (except age itself). 
We included an intercept for data-level mixed models, but not in the latent model as latent variables are centered around zero to follow the standard normal prior distribution.

We chose dense neural networks consisting of two hidden layers with $250$ and $100$ units and $\tanh$ activation function within encoder and decoder network architectures. We trained neural networks with the Adam optimizer~\cite{kingma2017adam} with standard hyperparameters and mixed models using the L-BFGS algorithm with $\eta=0.15$~\cite{liu1989limited}.
In each epoch we updated the VAE parameters $100$ times on the whole dataset before refitting the mixed model parameters. We trained for $20$ epochs, after which we observed no further improvement in total loss and reconstruction quality for more than two epochs.
The latent space dimensionality was set to $d=3$, chosen by domain knowledge to be expressive enough for three aspects of SMA. First, signal forwarding, meaning the short-term efficiency of neuromuscular transmission, second motor neuron integrity, reflecting the slowly varying capacity of the motor unit pool, and lastly the development of motor function.
In the loss function, we down-weighted the KL divergence loss term by $0.5$ because of the modest data dimensionality which led to an impact of reconstruction quality for higher KL-weighting. The weighted KL term converged to contribute roughly $10\%$ to the loss. The alignment terms were up-weighted by $5$ and converged to contribute about $2.5\%$ each to the final loss, placing primary emphasis on reconstruction quality. Model performance remained stable when these coefficients were halved or doubled. Increasing the alignment terms tended to increase skewness of the empirical test-statistic distribution, whereas decreasing them led to larger deviations between encoder outputs and latent mixed-model predictions. We chose a diagonal covariance structure for $\boldsymbol{\Phi},\boldsymbol{\Sigma}$. We found that alternative covariance structures did not contribute to a better reconstruction quality of test items, while conditional residuals were approximately uncorrelated. 

To simulate an artificial treatment effect, we first computed the incremental score to be added for each observation; no extra score was added to measurements before the treatment switch, after the switch an artificial improvement of one item every six months per test was applied. This increment was implemented by randomly selecting test items for each measurement instrument that had not yet reached their maximum value and increasing their score by one for each subsequent observation. If an item was already at its full score in one of these subsequent observations, the additional points were allocated to another randomly chosen item if possible.

\newpage
\section{Tables}
\label{tab}

\begin{table}[!ht]
\centering
\small
\begin{tabular}{p{3cm}p{1.6cm}p{3.6cm}p{3.9cm}}
\toprule
\textbf{Full Name} & \textbf{Abbr.} & \textbf{Test Description} & \textbf{Details} \\
\midrule
Revised Upper Limb Module\cite{RULM} & RULM & Measures upper limb function and strength across tasks like reaching, lifting, and hand movements. & 
\begin{tabular}[t]{@{}ll@{}}
\textbf{Target} & Children \\ 
\textbf{Median\,Age} & $11.1$ \\ 
\textbf{Items} & $20+1$ \\ 
\textbf{Max\,Score} & $37+6$ \\ 
\textbf{Inputs} & $63$ \\ 
\textbf{Obs.} & $3625$
\end{tabular} \\
\midrule
Hammersmith Functional Motor Scale Expanded\cite{HFMSE} & HFMSE & Assesses gross motor skills, including activities like sitting, rolling, standing, and transitional movements. & 
\begin{tabular}[t]{@{}ll@{}}
\textbf{Target} & Children \\ 
\textbf{Median\,Age} & $9.6$ \\ 
\textbf{Items} & $33+1$ \\ 
\textbf{Max\,Score} & $66+6$ \\ 
\textbf{Inputs} & $104$ \\ 
\textbf{Obs.} & $3279$
\end{tabular} \\
\midrule
Children's Hospital of Philadelphia Infant Test of Neuromuscular Disorders\cite{CHOP} & CHOP-INTEND & Assesses neuromuscular function, concentrating on spontaneous and prompted movements. & 
\begin{tabular}[t]{@{}ll@{}}
\textbf{Target} & Infants \\ 
\textbf{Median\,Age} & $3.0$ \\ 
\textbf{Items} & $16$ \\ 
\textbf{Max\,Score} & $64$ \\ 
\textbf{Inputs} & $68$ \\ 
\textbf{Obs.} & $1833$
\end{tabular} \\
\midrule
Hammersmith Infant Neurological Examination, Section 2\cite{HINE} & HINE-2 & 
Monitors early development, especially the attainment of motor function milestones (sitting, standing, walking).& 
\begin{tabular}[t]{@{}ll@{}}
\textbf{Target} & Infants \\ 
\textbf{Median\,Age} & $4.4$ \\ 
\textbf{Items} & $8+3$ \\ 
\textbf{Max\,Score} & $26+3$ \\ 
\textbf{Inputs} & $40$ \\ 
\textbf{Obs.} & $4071$
\end{tabular} \\
\midrule
ALS Functional Rating Scale Revised\cite{ALSFRS} & ALSFRS-R & Covers speech, swallowing, and fine motor tasks; designed for ALS but also used for SMA. & 
\begin{tabular}[t]{@{}ll@{}}
\textbf{Target} & Adults \\ 
\textbf{Median\,Age} & $34.9$ \\ 
\textbf{Items} & $12+1$ \\ 
\textbf{Max\,Score} & $48+4$ \\ 
\textbf{Inputs} & $65$ \\ 
\textbf{Obs.} & $732$
\end{tabular} \\
\bottomrule
\end{tabular}
\caption{Measurement instruments to assess a patient's motor function in SMArtCARE. Some tests contain additional items that are not used to determine the official test score. However, we still include these items to increase the amount of information.}
\label{tab:measurement}
\end{table}

\begin{table}[!ht]
\centering
\small
\begin{tabularx}{\textwidth}{ll}
\toprule
\textbf{Symbol} & \textbf{Meaning} \\
\midrule

$i\in\mathscr{I}$ & Patient index \\
$t\in\mathscr{T}_i, t\in\mathscr{T}_{i,l}$ & Observation times\\
$l\in\mathscr{L}, l\in\mathscr{L}_{i,t}$ & Measurement instrument index\\
$b\in\mathscr{B}$ & Bootstrap samples \\
$n, n_l$ & Data dimension \\
$m_i, m_{i,l}$ & Number of observations \\
$d$ & Latent dimension \\
$p,q$ & Number of fixed and random effect coefficients \\
$rd$ & Difference in parameters between nested models \\
$Y, \mathbf{y}_i / Z, \mathbf{z}_i$ &  Single data/latent observation \\
$Y_i, \mathbf{Y}_i=\big(\mathbf{y}_{i,t}\big)^\top_{t\in\mathscr{T}_i}$ & Data trajectory\\
$Z_i, \mathbf{Z}_i=\big(\mathbf{z}_{i,t}\big)^\top_{t\in\mathscr{T}_i}$ & (Latent) response \\
$Y_{i,l}, \mathbf{Y}_{i,l}=\big(\mathbf{y}_{i,t,l}\big)^\top_{t\in\mathscr{T}_{i,l}}$ & Instrument specific data trajectory\\
$Z_{i,l}, \mathbf{Z}_{i,l}=\big(\mathbf{z}_{i,t,l}\big)^\top_{t\in\mathscr{T}_{i,l}}$ & Instrument specific latent trajectory \\
$\widehat{\mathbf{Z}}_i=\big(\widehat{\mathbf{z}}_{i,t}\big)^\top_{t\in\mathscr{T}_i}$ & Latent mixed model estimation \\
$\mathbf{X}_i, \mathbf{T}_i$ & Design matrices for fixed and random effects\\
$\mathbf{B},\widehat{\mathbf{B}}$& Fixed effect coefficient matrix \\
$U_i,\mathbf{U}_i,\widehat{\mathbf{U}}_i$ & Random effects \\
$E_i,\mathbf{E}_i$ & Residual error \\
$\boldsymbol{\Phi}, \boldsymbol{\Sigma},\mathbf{V}_i$ & Random effects covariance, residual covariance, precision  \\
$\mathbf{I}_d$ & Identity matrix of dimension $d$\\
$P_{Y,Z},p_{Y,Z}(\mathbf{y},\mathbf{z})$ & Joint generative distribution of $(Y,Z)$ \\
$P_Z,p_{Z}(\mathbf{z})$ & Prior on latent variable \\
$P_{Y\mid \mathbf{z}},p_{Y\mid \mathbf{z}}(\mathbf{y})$ & Conditional sampling distribution of $Y$ given latent $\mathbf{z}$ \\
$P_Y,p_{Y}(\mathbf{y})$ & Marginal distribution of $Y$ \\
$P_{Z\mid \mathbf{y}},p_{Z\mid \mathbf{y}}(\mathbf{z})$ & True posterior of $Z$ given $\mathbf{y}$ \\
$Q_{Z\mid \mathbf{y}},q_{Z\mid \mathbf{y}}(\mathbf{z})$ & Variational posterior of $Z$ given $\mathbf{y}$ \\
$\mathrm{enc}_{\boldsymbol{\phi}}, \mathrm{enc}_{\boldsymbol{\phi}_l}, \mathrm{dec}_{\boldsymbol{\theta}}, \mathrm{dec}_{\boldsymbol{\theta}_l}$ & Encoder/decoder neural network \\
$\mathcal{L}_{\mathrm{VAE}}$ & negative ELBO loss for VAE \\
$\mathcal{L}_{\mathrm{ML}},\mathcal{L}_{\mathrm{REML}}$ & (Restricted) ML log-likelihood \\
$\mathcal{L}_{\mathrm{MMVAE}}$ & Joint loss for multi-measurement VAE + mixed model \\
$\beta, \eta,\gamma$ & Weighting of KL and alignment terms in $\mathcal{L}_{\mathrm{MMVAE}}$ \\
\bottomrule
\end{tabularx}
\caption{Notation used throughout the manuscript.}
\label{tab:notation}
\end{table}

\begin{table}[!ht]
\centering
\begin{tabular}{@{}lccccc@{}}
\toprule
& \multicolumn{3}{c}{Real} & \multicolumn{2}{c}{Artificial $+2$} \\
\cmidrule(lr){2-4}\cmidrule(lr){5-6}
Instrument & Mean & SD & Percentage & Mean & SD  \\
\midrule
HINE-2   & $0.87$ & $0.20$ & $3.00$ & $2.84\ (+1.97)$ & $0.57$ \\
RULM     & $1.47$ & $0.12$ & $5.09$ & $2.62\ (+1.15)$ & $0.41$ \\
CHOP     & $0.91$ & $0.31$ & $3.17$ & $2.64\ (+1.73)$ & $0.46$ \\
HFMSE    & $1.56$ & $0.38$ & $5.37$ & $2.72\ (+1.16)$ & $0.59$ \\
ALSFRS-R & $0.52$ & $0.30$ & $1.81$ & $1.30\ (+0.78)$ & $0.60$ \\
\bottomrule
\end{tabular}
\caption{Mean predicted item-level differences for five motor function instruments (averaged over all patients and ten random seeds) between predictions with and without treatment switch one year after the switch took place. Results are based on a three-dimensional latent mixed effects trajectory. Positive values indicate improved motor function compared to no switch. The SD column lists the standard deviation of patient-level mean values across the ten random seeds. The percentage column converts absolute differences to a percentage of the instrument’s maximum sum-score to enable cross-instrument comparability. The left block refers to the switch detected in the unaltered data; the right to an artificial switch of $+2$ items per year. Here, changes in mean values w.r.t. unaltered trajectories are indicated in brackets.}
\label{tab:data_level_effect}
\end{table}

\end{document}